# Towards Santali Linguistic Inclusion: Building the First Santali-to-English Translation Model using mT5 Transformer and Data Augmentation

SYED MOHAMMED MOSTAQUE BILLAH*, ATEYA AHMED SUBARNA *, SUDIPTA NANDI SARNA, AHMAD SHAWKAT WASIT, ANIKA FARIHA

BRAC University, Bangladesh

ASIF SUSHMIT

Bengali.AI, Bangladesh

Dr. FARIG YOUSUF SADEQUE

BRAC University, Bangladesh

Around seven million individuals in India, Bangladesh, Bhutan, and Nepal speak Santali, positioning it as nearly the third most commonly used Austroasiatic language. Despite its prominence among the Austroasiatic language family's Munda subfamily, Santali lacks global recognition. Currently, no translation models exist for the Santali language. Our paper aims to examine the feasibility of building Santali translation models based on available Santali corpora. The paper successfully addressed the low-resource problem and, with promising results, examined the possibility of creating a functional Santali machine translation model in a low-resource setup. Our study shows that Santali-English parallel corpus performs better when in transformers like mt5 as opposed to untrained transformers, proving that transfer learning can be a viable technique that works with Santali language. Besides the mT5 transformer, Santali-English performs better than Santali-Bangla parallel corpus as the mT5 has been trained in way more English data than Bangla data. Lastly, our study shows that with data augmentation, our model performs better.

## 1 INTRODUCTION

With over seven million native speakers from India, Bangladesh, Bhutan, and Nepal, the third most widely used Austroasiatic language is Santali. Since Santali had no written language before the nineteenth century, all shared information was transmitted verbally from one generation to the next. Before 1860, European missionaries, folklorists, and anthropologists like A. R. Campbell, Lars Skrefsrud, and Paul Bodding had worked on transcribing Santali using Bengali, Odia, and Roman scripts. As a result of these efforts, we now have Santali dictionaries, adaptations of folktales, and studies on the language's morphology, syntax, and phonetic structure. Although Santali holds the distinction of being the predominant language within the Munda subfamily of Austroasiatic languages, it has not garnered significant global recognition, resulting in its classification as a low-resource language. Like numerous other languages, Santali has received no attention in modern NLP research, which primarily concentrates on approximately 20 out of the 7000 languages spoken globally. As a consequence, Santali remains largely unstudied in the field of NLP. Low-resource languages are primarily those that lack extensive monolingual or parallel corpora, in addition to the manually generated language resources required for creating statistical NLP applications. NMT, or Neural Machine Translation relies heavily on vast amounts of data for accurate translations, it is a data-intensive technology. These models are built with neural networks, which are capable of finding intricate patterns in data. However, their effectiveness is closely tied to the quantity as well as the quality of the data that is trained. In the case of a lack of data, NMT models may struggle to identify patterns and produce flawed translations. The absence of such a corpus has a negative impact on the models' performance. In other words, NMT systems may fail to attain desirable results for low resource languages. Since Santali is a low-resource language spoken by a minor group of people, getting a parallel corpus is quite challenging. That is why our goal is to find appropriate machine translation algorithms and tools for dealing with low resource languages, such as data augmentation and transfer learning, in order to achieve better outcomes. There was a time when a substantial amount of data in the form of parallel corpus was necessary to train a translation model. However, with the advent of newer transformer models, the reliance on data has diminished somewhat. Consequently, we wanted to assess whether the available internet resources would suffice to create a translation model for the Santali language, which plays a crucial role in the communication of a community. This thesis allows us to gauge the potential for work on a language with limited resources. If we achieve promising results in this domain, we plan to seek monetary funds from government or private organizations to advance this approach. In short, our objective is to determine if the current conditions are favorable for working with the low-resource Santali language in the field of machine translation. Furthermore, we aspire for our work to serve as the gateway to the realm of Santali machine translation.

---

* Both authors contributed equally to the research.

## 2 RELATED WORKS

The Encoder Decoder Neural Machine Translation [1] had been showing promising results, but only on large data. When it comes to the low resource language, NMT fails to show any effective performance. This paper [2] proposes a Transfer model for low resource NMT, that significantly improves performance. Their proposed method is to primarily train the model with a high resource language pair and then pass the obtained parameter to the low resourced pair. They are calling the first one a parent model and a second one a child model. With this idea, they were able to get an average of 5.6 BLEU on four language pairs with few resources. Transfer learning is basically transferring the knowledge from a learned task to any related task. Here it passes the learned parameters. As a result, the child task needs less training data. Firstly training a NMT model as the parent with sufficient data, then training another NMT model (**child model**) initializing with the parameters of the parent model. They also focused on the choice of parent language. The paper depicts a comparison between French and German as parent language for Spanish language, where French performs better than German due to the similarity of French and Spanish. Extending this knowledge, they further carried experiments on relatedness of parent class. The results indicates, performance of such models increases for "*closely related*" parent classes. As a result, Training with transformers significantly improves translation score, as transformers already have previous context. That's why it is one of the most crucial elements while working with low-resource languages. This paper [3] suggests that inner layers of a model's network play a critical role than the embeddings. Moreover, transfer learning speeds up training by nullifying the need for warm up stages. A warm up stage is when initially the learning rate is gradually increased to stabilize the training process and reduce the impact of noisy gradients.

Additionally, to address the low resource problem of a parallel corpus, data augmentation can be a technique proposed by this paper [4]. Data augmentation is increasing parallel corpora by adding synthetic data. It is largely inspired from computer visioning models. A picture of a boy playing baseball and the mirror image of the same picture would have the same labeling which would increase the dataset. Similarly, the paper targets low frequency words in the dataset to create new sentences synthetically to improve their parameter estimation. The paper shows that using data augmentation the BLEU score has been improved up to by 2.9 points and up to 3.2 points in case of back-translation score.

## 3 DATA COLLECTION, PREPROCESSING AND DATA ANALYSIS

### 3.1 DATA COLLECTION

The data is taken from a website called "bible.com", and it is currently the biggest dataset that is available all over the internet as the Bangladeshi-Santali data set. The Santali Bible had 66 books and multiple chapters in each book. Among them "SEREKO 1" had the highest number of chapters (150). There were several books with only one chapter. We built a customized data crawler to crawl through all the books and all the chapters to fetch all the Santali-English-Bengali parallax corpora that were available for us. From there, we found 29,651 sections, which may or may not contain more than one sentence. The total number of sentences was 69,086; the total number of words was 6,63,684. The length of the word set was 50,822. The top 10 most frequently occurring words are as follows:

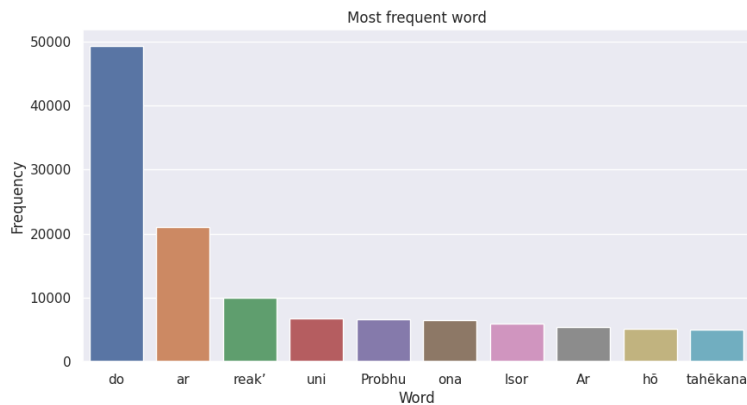

Figure 1: Barplot of most frequent words (top 10)

Since the Santali Bible database does not contain enough information. Most of the words from the unique word list did not occur more than once. For example, more than half of the words found were present only once. Which is definitely not enough for any model to learn its context. The number of words that occurred only once is 28,525; the number that occurred only twice is 7,142. Here is the list of the top 10 frequencies of the least frequently occurring words in the whole corpus:

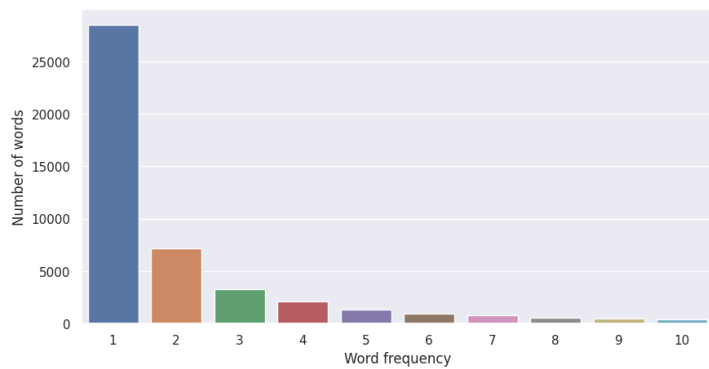

Figure 2 :Bar plot of frequency of least occurring words. (Top 10)

## 3.2 DATA PREPROCESSING

*3.2.1 Data Cleaning:*

After data extraction, we executed some basic pre-processing techniques, such as the removal of punctuation marks. Then we had to encode the Santali data since it uses special Roman characters. Besides, we had to do some sentence level assignments which had been discussed in data splitting section. The data was very clean to begin with, so we did not need to go through rigorous pre-processing.

*3.2.2 Data Splitting and Sentence Alignment:*

Our dataset included sets of sentences as one unit pair where two or more sentences were grouped as one sentence having a group of corresponding target sentences, which may or may not be of the same length. For example,

| Santali | English |
|---------|---------|
| 1)Isore menkeda, "Sermare marsalak'ko hoyok'ma, ar onako do ńińdạ khon sińe begarma. Onako do eạak' eạk' dinko, candoko ar bochorko reak' cinhạko tahenma. | 1)And God said, "Let there be lights in the expanse of the heavens to separate the day from the night. And let them be for signs and for seasons, and for days and years. |
| 2) Isor do dhạrtiren sanam lekan bir janwarko, ạsul janwarko ar sanam lekan leńok' ṭuṇḍạṅkan jiwiyankoe benaoket'koa. Noko sanamkoge apan ạpin jạtrenko lekate saṅgek' reạk' dạeye emat'koa. Isore ńelkeda ona do ạdi moṅgge hoyakana. | 2)And God made the beasts of the earth according to their kinds and the livestock according to their kinds, and everything that creeps on the ground according to its kind. And God saw that it was good. |

Figure 3 : Examples of sentences in the Dataset

We see in example-1 the source unit and the target unit have the same number of sentences, while in example-2, the sentence number varies in the source and target unit. Such lengthy units hardly can perform well in NMT models. So, we broke down the dataset in two parts-namely, 'same length units', 'variable length units'. For the same length units, we broke-down the units into single sentences and mapped each source sentence to the target sentence. And, on the variable length units, we kept the dataset unchanged. Lastly, to ensure our training corpus has both one-to-one mapped sentences and variable length units, we split the same length units and variable length units in 0.8,0.1.0.1 for the train, test and validation set and merged them in the final training, testing and validation data.

Algorithm followed in creating the dataset –

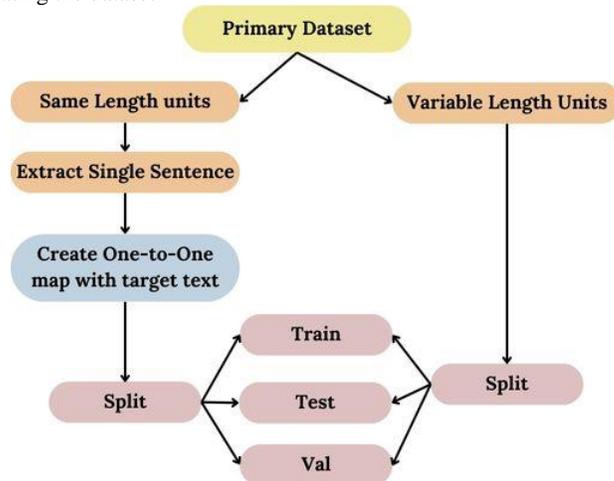

Figure 4: Flowchart to create train-test-validation dataset

Following the algorithm we get –

Table 1 : One-to-One mapping of Source and Target sentences

| Dataset | Primary Text Units | One to one sentence | Variable length units | Final text units |
|---|---|---|---|---|
| **Santali-English** | 29,651 | 27,295 | 10,950 | 38,245 |
| **Santali-Bangla** | 29,166 | 22,764 | 15,452 | 38,216 |

Train- Test-Validation split of the dataset–

Table 2 : Train-Test-Validation split for Santali-English Dataset

| Santali English | One-to-One Sentences | Variable Length Units | Total |
|---|---|---|---|
| *Train* | 21,836 | 8,760 | 30,596 |
| *Test* | 2,729 | 1,095 | 3,824 |
| *Validation* | 2,729 | 1,095 | 3,824 |

Table 3: Train-Test-Validation split for Santali-Bangla Dataset

| Santali Bangla | One-to-One Sentences | Variable Length Units | Total |
|---|---|---|---|
| Train | 18,211 | 12,361 | 30,572 |
| Test | 2,276 | 1,545 | 3,821 |
| Validation | 2,276 | 1,545 | 3,821 |

## 3.3 DATA ANALYSIS

### 3.3.1 Word2Vec

Then comes the vectorization of the words. Meaning that words needed to be presented through numerical value in a way that they held some contextual information. The creation of this word embedding was also crucial for us, as from this we could understand whether our data is able to preserve any contextual quality or not. The results were optimistic, as when we wanted to find the most similar words to "Ishak," the prophet, it gave us ten more names who were likely prophets themselves:

```
1 model.wv.most_similar(positive=["Isahak"])

[('Abraham', 0.9850263595581055),
 ('Johan', 0.9794796705245972),
 ('Eliọ', 0.9743596911430359),
 ('Jakob', 0.9739243984222412),
 ('Amasa', 0.972024261951465),
 ('Joab', 0.97163218259814),
 ('Asa', 0.9692220687866211),
 ('Ahitophel', 0.9671008586883545),
 ('kedeteye', 0.9667977690696716),
 ('Basa', 0.9656971096992493)]
```

Figure 5: Most similar words to "Ishahak" based on cosine similarity. (Top 10)

Our model can also produce a list of most similar words, and it can solve equations to give us words. Such as king + woman + man = queen. But unfortunately, since we do not possess much knowledge of the language, we cannot show this task in Santali in a way that will be understandable to us. Our model can also produce a lower-dimensional representation of words with their cosine similarity. Such a graph is shared below that contains ten most similar and dissimilar words. Graph for "Isor" is shared bellow:

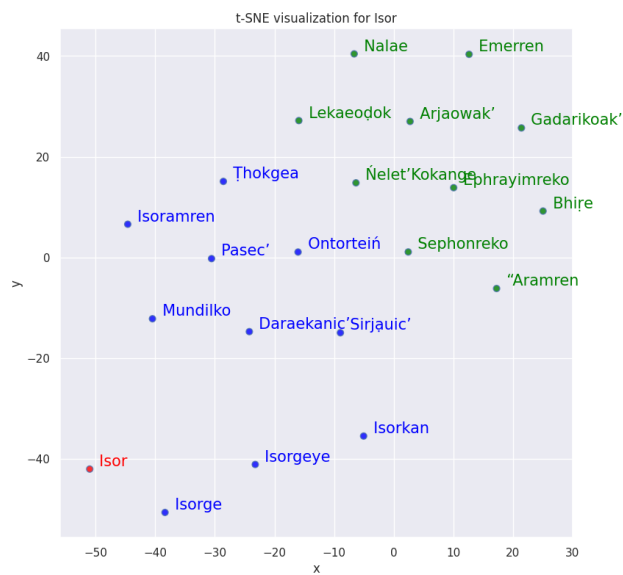

Figure 6: t-SNE plot to visualize word vectors in a 2D plain. Red- the source word, Blue - similar words, green- dissimilar words.

### 3.3.2 Tokenizer

MT5 model is compatible with only sentence piece tokenizers.

For our Santali-English dataset, we extracted the Santali text from the dataset, and loaded the T5 tokenizer and trained it on our Santali dataset, took the top 32,000 word pieces as tokens which is here our source text. And for the English part, we used the t5-small tokenizer from "Autotokenizer" and tokenized out data.

After Tokenizing a sentence like - **"re isor do ot ar sermae sirjaukeda"**

It becomes -

$$[\text{" re", " isor", " do", " ot", " ar", " sermae", " sirjau", " keda"}]$$

Here, each word is assigned to a corresponding number.

For the Santali-Bangla dataset, similarly first we extracted the Santali and Bangla text. We used the previously trained Santali tokenizer to tokenize our source text. And for the Bangla text which is our target text, we used the pre-trained Bangla tokenizer[1] .

## 4   METHODOLOGY

### 4.1   DATA AUGMENTATION

#### 4.1.1 Santali-English Data Augmentation:

For our Santali to English dataset, we augmented the English part of our training corpus. Our primary training dataset contained 26,724 pairs of sets of sentences, where 20k pairs were one-to-one mapped. We labeled this as a 'good dataset'

---

1 https://huggingface.co/csebuetnlp/banglat5

and augmented these 20k sentences. We excluded the sentences, which were not one-to-one mapped, from the augmentation process, since the sentences were too long, and making tiny changes to such long sentences would not affect much to create comparatively newer sentences. From the "TextAttack" [5] framework we used the "EasyDataAugmenter" class as our augmentation recipe. This class implements-

- WordNet synonym replacement: Here a word is selected randomly and then replaced with a synonym
- Word deletion: Removing a word randomly.
- Word order swaps: Randomly choose two words and swap their position
- Random synonym insertion: Select a word randomly and insert it's synonym in a random position

All this is done in one Augmentation method. After augmentation, we had a training corpus of 46,726 sets of pairs of sentences.

*4.1.2 Santali-Bangla Data Augmentation:*

In order to augment the Bangla texts we used bnaug[2] library. Similar to the English text, we extracted the Bangla texts from the pairs and passed it to the "bnaug" library. We used token replacement and back translation both on each Bangla sentences to generate a single augmented instance. For token replacement we utilized random mask-based generation using pre-trained Bengali GloVe10 and Word2Vec11 embeddings. Again, using the library's back translation method we paraphrased the mask based augmented sentences to paraphrase. Initially, in our training corpus we had 27,060 sets of pairs of sentences, however with augmentation, we generated more 15,000 pairs of sentences. Finally, our training corpus had 42,060 sets of pairs of sentences.

The algorithm we followed, creating the augmented dataset for both Bangla and English:

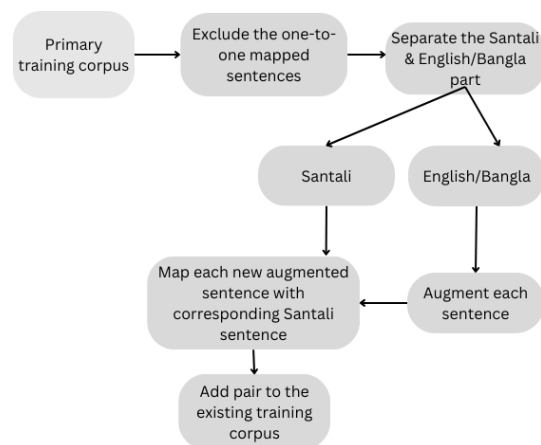

Figure 7: Data Augmentation Flowchart

## 4.2 SEQ2SEQ MODEL

As proposed in this paper [6] we implement this model with two Recurrent Neural Networks. The first RNN is used to create the encoder part and using the second RNN we created the decoder part. The encoder part had an embedding layer and then a GRU layer. In the encoder, the forward method takes in an input tensor and a hidden state tensor as input. Then

---

the input tensor as embedded passes through the GRU, to obtain the output tensor, which gives the hidden state of the current time step. For the decoder, we created an Attention based [7] decoder RNN. The RNN consists of six layers: Embedding Layer, Attention layer, Attention Combination Layer, Dropout Layer, GRU layer, and Output Layer. We pass the hidden states, output language unique word size, and dropout probability to generate the next token. The forward method takes the input tensor, the hidden state tensor, and the encoder output tensor as the input. Then it embeds the input tensors and applies the attention mechanism to the encoder output tensor. Then in the GRU layer, the combined attention vector passes through and updates the hidden states. The attention mechanism helps the decoder to focus on different parts of the input sequence during the decoding process. The decoder returns hidden states and output tensor and attention weights. We initialized the hidden size as 256 for the encoder and the attention decoder. We use the 'topk' function to select the most probable next token to generate our sequence in a probability distribution.

### 4.3 MT5-SMALL

In order to leverage transfer learning [2], we finetune MT5-small [8] checkpoints which were trained on text and code of 101 different languages.It has only 300M parameters compared to 117 billion parameters of t5 model, making it a suitable choice to finetune on a smaller dataset. The model follows the architecture of T5 [9] which is again a vanilla encoder decoder model [7].

- Encoder: The encoder consists of 6 encoder layers, each with 12 attention heads. The attention heads are responsible for attending to different parts of the input sequence.
- Decoder: The decoder consists of 6 decoder layers, each with 12 attention heads. The decoder generates the output sequence one token at a time, by attending to the previous tokens in the sequence and the encoder output.

## 5 RESULT ANALYSIS

First, we try the seq2seq model vs mt5 model to compare their ability to translate Santali to English.

Table 4: BLEU 4 score comparison between Seq2Seq and mt5 model

| Model Type | Validation Score (BLEU 4) | Test Score (BLEU 4) |
|---|---|---|
| Seq2Seq | 3.33 | 1.87 |
| mT5(Santali-English) | 8.93 | 6.79 |

Due to attributes of transfer-learning, mT5 works better than Seq2Seq [2, 8]  while dealing with English sentences as mt5 has prior context about how English as a language works. Since this proves that mT5 works better than seq2seq, we won't train further seq2seq models to save resources.

Now, we will discover whether data augmentation works better for Santali-English or not.

Table 5: BLEU 4 score comparison with data augmentation technique (Santali-English)

| Model Type | Validation Score (BLEU 4) | Test Score (BLEU 4) |
|---|---|---|
| *mT5 with data augmentation (Santali-English)* | 11.13 | 10.5 |
| *mT5 without data augmentation (Santali-English)* | 8.93 | 6.79 |

Data augmentation clearly improved the result of Santali-English translation. This proves that data augmentation is a technique that will work while working with Santali to English translation.

Table 6 :BLEU 4 score comparison with data augmentation between "Santali-English" and "Santali-Bangla" models

| Model Type | Validation Score (BLEU 4) | Test Score (BLEU 4) |
|---|---|---|
| *mT5 with augmentation (Santali-English)* | 11.13 | 10.5 |
| *mT5 with augmentation (Santali-Bangla)* | 2.85 | 1.59 |

Bangla performs poorly because English tokens were six times higher [8] then Bangla tokens in the mC4 dataset that was used while training mt5. So, it can be deduced that performance enhancement of models due to transfer-learning has direct relation to the tokens of languages that it had been trained on.

## 6 LIMITATIONS

### 6.1 RESOURCES

Due to limited computational units, we could not meticulously test every test case. For example, when we deduced that mt5 small works better than Seq2Seq, tested only in case of Santali to English translation. We could not test reserves translation, translation with augmented data, Santali to Bangla translation or Bangla to Santali translation. We did it to reduce our model training cost.

### 6.2 LANGUAGE EXPERTISE

Since our team does not possess expertise on Santali language, we could not synthetically increase Santali data.

### 6.3 DATASET

Santali has two alphabet styles – "Ol-chiki" and roman. Indian Santali mostly use the "Ol-chiki" version of Santali. Since due to shortage of lingual expertise, we could not establish a bridge between "Ol-chiki" to roman Santali, we could not use wikipedia-scrapped "Ol-chiki" dataset, resulting in a shorter dataset.

## 6.4 MODEL LACKS OUT OF DOMAIN KNOWLEDGE

Since the model is being only trained on biblical data, it lacks knowledge on conversational data. Due to lack of knowledge from other domains and writing style, our trained model cannot be used in real-life translation tools.

## 7 CONCLUSION

In the era of globalization, the presence of translation models can elevate a particular language-speaking population to a global stage, diminishing the language barrier. Unfortunately, not all languages have their own translation model. Santali, a language spoken by seven million people across multiple countries in the world, is such a language. The main barrier to creating a Santali translation model is the shortage of data. But in the past few years, the field of NLP has crossed exemplary milestones and given models that can work even with very little data. Our primary objective was to determine the feasibility of constructing a translation model for the Santali language by exploring the limited online resources. Although the available internet resources were scarce, our adoption of advanced techniques such as transfer learning, augmented datasets, and transformer models yielded a commendable BLEU score. This success has instilled confidence in us that we can develop a practical Santali language translation model with access to a substantial dataset.